\documentclass{article}

\usepackage[preprint]{neurips_2022}

\usepackage[utf8]{inputenc} 
\usepackage[T1]{fontenc}    
\usepackage{hyperref}       
\usepackage{url}            
\usepackage{booktabs}       
\usepackage{amsfonts}       
\usepackage{nicefrac}       
\usepackage{microtype}      
\usepackage{xcolor}         

\usepackage{wrapfig}
\usepackage{graphicx}

\usepackage{amsmath,amsfonts,bm}

\def\eqref#1{equation~\ref{#1}}

\def\1{\bm{1}}

\DeclareMathAlphabet{\mathsfit}{\encodingdefault}{\sfdefault}{m}{sl}
\SetMathAlphabet{\mathsfit}{bold}{\encodingdefault}{\sfdefault}{bx}{n}
\newcommand{\tens}[1]{\bm{\mathsfit{#1}}}

\def\tE{{\tens{E}}}

\def\tX{{\tens{X}}}

\usepackage{cleveref}
\usepackage[ruled,vlined]{algorithm2e}
\usepackage{caption}
\usepackage{subcaption}
\usepackage{mathtools}
\usepackage[figuresleft]{rotating}

\definecolor{mycolor}{RGB}{83,83,182}
\hypersetup{
    colorlinks=true,
    citecolor=mycolor,
    linkcolor=mycolor
}

\title{PAVI:\\Plate-Amortized Variational Inference}

\author{%
    Louis Rouillard\\
    Université Paris-Saclay, Inria, CEA\\
    Palaiseau, 91120, France\\
    \texttt{louis.rouillard-odera@inria.fr}
    \And
    Thomas Moreau\\
    Université Paris-Saclay, Inria, CEA\\
    Palaiseau, 91120, France\\
    \texttt{thomas.moreau@inria.fr}
    \And
    Demian Wassermann\\
    Université Paris-Saclay, Inria, CEA\\
    Palaiseau, 91120, France\\
    \texttt{demian.wassermann@inria.fr}
}

\begin{document}

\maketitle

\begin{abstract}
  Given some observed data and a probabilistic generative model, Bayesian inference aims at obtaining the distribution of a model's latent parameters that could have yielded the data.
This task is challenging for large population studies where thousands of measurements are performed over a cohort of hundreds of subjects, resulting in a massive latent parameter space.
This large cardinality renders off-the-shelf Variational Inference (VI) computationally impractical.
In this work, we design structured VI families that can efficiently tackle large population studies.
To this end, our main idea is to share the parameterization and learning across the different i.i.d. variables in a generative model -symbolized by the model's \textit{plates}.
We name this concept \textit{plate amortization}, and illustrate the powerful synergies it entitles, resulting in expressive, parsimoniously parameterized and orders of magnitude faster to train large scale hierarchical variational distributions.
We illustrate the practical utility of PAVI through a challenging Neuroimaging example featuring a million latent parameters, demonstrating a significant step towards scalable and expressive Variational Inference.
\end{abstract}

\section{Introduction}
Population studies correspond to the analysis of measurements over large cohorts of human subjects.
These studies are ubiquitous in health care \citep{fayaz2016prevalence,towsley2011population}, and can typically involve hundreds of subjects and thousands of measurements per subject.
For instance in the context of Neuroimaging~\citep{kong_spatial_2018}, measurements $X$ can correspond to signals measured in hundreds of locations in the brain for a thousand subjects.
Given this observed data $X$, and a generative model that can produce data given some model parameters $\Theta$, we want to recover the latent $\Theta$ that could have yielded the observed $X$.
In our Neuroimaging example, $\Theta$ can be local labels for each location and subject, together with global parameters common to all subjects --such as the brain connectivity corresponding to each label.
We are interested in recovering the \textit{distribution} of the $\Theta$ that could have produced $X$.
Following the Bayesian inference formalism~\citep{Gelman_book}, we cast both $\Theta$ and $X$ as sets of Random Variables (RVs) and our goal is to recover the \textit{posterior} distribution: $p(\Theta|X)$.
Due to the nested structure of the considered applications we will focus on the case where $p$ corresponds to a Hierarchical Bayesian Model (HBM) \citep{Gelman_book}.
In the particular context of population studies, the multitude of subjects and measurements per subject implies a large dimensionality for both $\Theta$ and $X$.
This large dimensionality in turn creates computational hurdles that we wish to overcome through our method.

To tackle Bayesian inference, several methods have been proposed in the literature.
Earliest works resorted Markov Chain Monte Carlo \citep{koller_probabilistic_2009}, which tend to be slow in high dimensional settings \citep{blei_variational_2017}.
Recent approaches, coined Variational Inference (VI), cast the inference as an optimization problem~\citep{blei_variational_2017, zhang_advances_2019}.
Within this framework, inference reduces to finding the parametric distribution $q(\Theta; \phi) \in \mathcal{Q}$ closest to the unknown posterior $p(\Theta | X)$ in a variational family $\mathcal{Q}$ chosen by the experimenter.
In recent years, VI has benefited from the advent of automatic differentiation \citep{ADVI} and the automatic derivation of the variational family $\mathcal{Q}$ based on the structure of the HBM $p$ \citep{ASVI, CF}.

To achieve competitive inference quality, VI requires the variational family $\mathcal{Q}$ to contain distributions closely approximating $p(\Theta | X)$ \citep{blei_variational_2017}.
Yet the form of $p(\Theta | X)$ is usually unknown to the experimenter.
To forgo a lengthy search for a valid family, one can instead resort to universal density approximators, such as normalizing flows~\citep{papamakarios_normalizing_2019}.
To achieve this generality, normalizing flows are highly parameterized and consequently scale poorly with the dimensionality of $\Theta$.
In large populations studies, as this dimensionality grows to the million, the parameterization of normalizing flows can in turn become prohibitively large.
This creates a detrimental trade-off between expressivity and scalability~\citep{ADAVI}.
To tackle this challenge, \citet{ADAVI} recently proposed --in the ADAVI architecture-- to partially share the parameterization of normalizing flows across the hierarchies of a generative model.
ADAVI had several limitations we improve upon in this work: removing the Mean Field approximation \citep{blei_variational_2017}; treating arbitrary HBMs instead of pyramidal HBMs only; and introducing non-sample-amortized variants.
Critically, while ADAVI tackled the over-parameterization of VI in population studies, it still could not perform inference in very large data regimes due to computational limits.
Indeed, as the size of $\Theta$ increases, the evaluation of a single gradient over the entirety of the architecture's weights quickly required too much memory and compute.
To overcome this second challenge, stochastic VI~\citep{SVI} subsamples the parameters $\Theta$ inferred for at each optimization step.
However, using SVI, the weights for the posterior of a given local parameter $\theta \in \Theta$ are only updated when $\theta$ is visited by the algorithm.
In the presence of hundreds of thousands of such local parameters, stochastic VI can become prohibitively slow.

In this work, we introduce the concept of \textit{plate amortization} (PAVI) for fast and universal inference in large scale HBMs.
Instead of considering the inference over local parameters $\theta$ as separate problems, our main idea is to share both the parameterization and learning across those local parameters --or equivalently across a model's \textit{plates}.
We first propose an algorithm to automatically derive an expressive yet parsimoniously-parameterized variational family from a plate-enriched HBM.
We then propose a hierarchical stochastic optimization scheme to train this architecture efficiently, obtaining orders of magnitude faster convergence.
Leveraging the repeated structure of plate-enriched HBMs, PAVI is able to perform inference over arbitrarily large population studies, with constant parameterization and training time as the cardinality of the problem augments.
We illustrate this by applying PAVI to a challenging human brain cortex parcellation, featuring inference of a million parameters over a cohort of $1000$ subjects, demonstrating a significant step towards scalable, expressive and fast VI.
\section{PAVI architecture}
\begin{wrapfigure}{R}{0.4\textwidth}
    \vspace{-15pt}
    \begin{center}
        \includegraphics[width=0.4\textwidth]{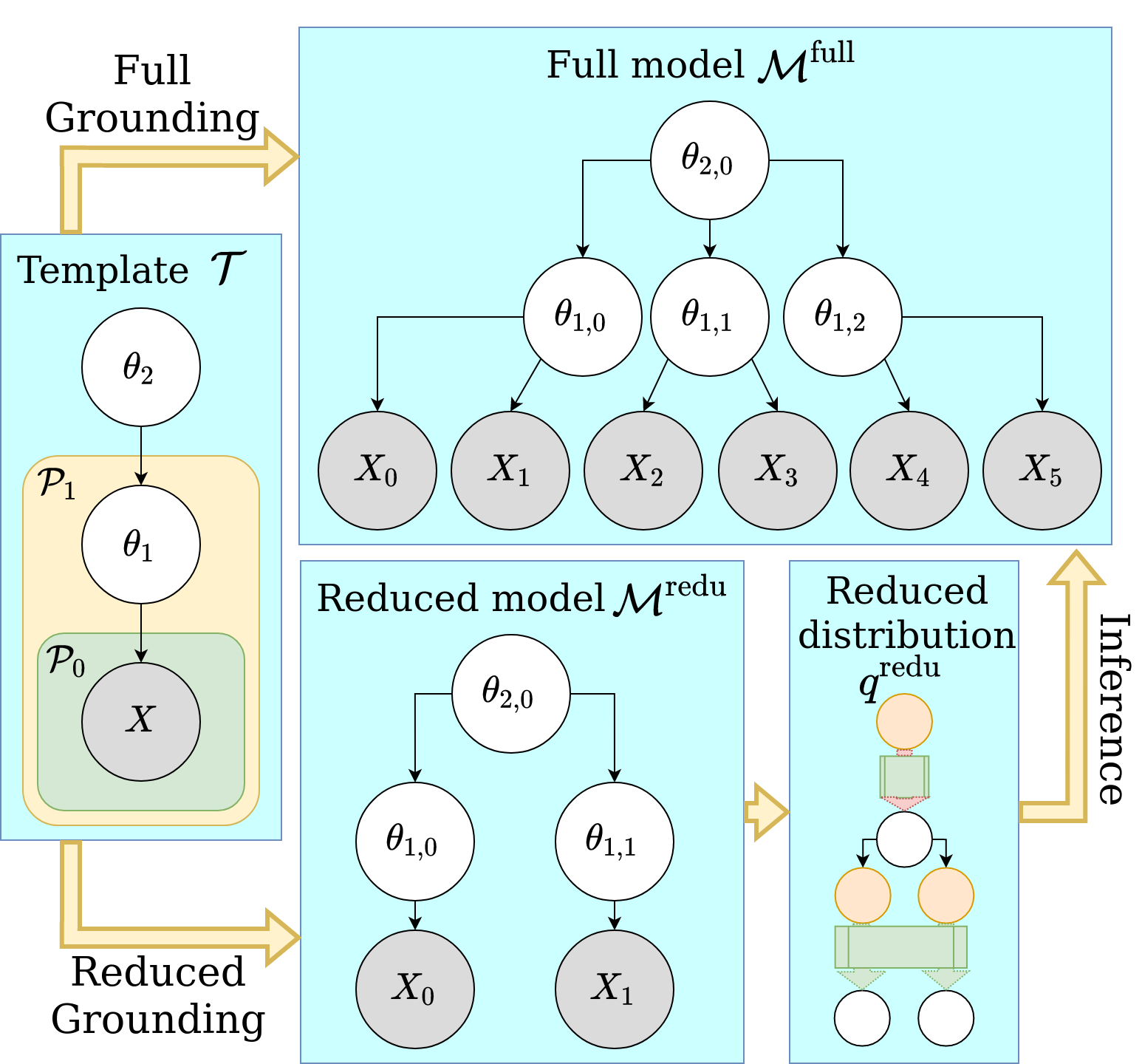}
    \end{center}
    \caption{\textbf{Plate Amortized Variational Inference (PAVI) working principle}.
    Starting on the left, the graph template $\mathcal{T}$ is grounded into 2 separate HBMs: $\mathcal{M^\text{full}}$ (top) and $\mathcal{M^\text{redu}}$ (down) of respective plate cardinalities $(3, 2)$ --large-- and $(2, 1)$ --small.
    Based on $\mathcal{M^\text{redu}}$, the reduced distribution $q^{\text{redu}}$ is constructed.
    We train $q^{\text{redu}}$ over data slices of small cardinality, before performing inference over the full model of large cardinality.}
    \label{fig:PAVI_principle}
    \vspace{-55pt}
\end{wrapfigure}
\label{sec:archi}
\subsection{Hierarchical Bayesian Models (HBMs), templates and plates}
\label{sec:HBMs}
Our objective is to perform inference in the context of large population studies modelled using plate-enriched Hierarchical Bayesian Models (HBMs).
These HBMs feature conditionally i.i.d. samples from a common conditional distribution at multiple levels, translating the graphical notion of \textit{plates} \citep{gilks_language_1994}.
\cref{fig:PAVI_principle} displays 2 toy instances of this i.i.d sampling: in our target applications, the total number of samples approaches the million \citep{kong_spatial_2018}. 

HBMs can be compactly represented via a Directed Acyclic Graphs (DAG) template $\mathcal{T}$ \citep{koller_probabilistic_2009} with vertices --corresponding to RV templates-- $X$ and $\Theta=\{ \theta_i \}_{i=1 .. I}$ and plates $\{ \mathcal{P}_p \}_{p=0 .. P}$.
We denote as $\operatorname{Plates}(\theta_i)$ the set of plates a given RV template $\theta_i$ belongs to.
$X$ corresponds to the sets of RVs observed during inference, and $\Theta$ to the parameters we want to infer.
Our goal is therefore to approximate the posterior distribution $p(\Theta | X)$.
In \cref{fig:PAVI_principle}, there are 2 latent RV templates: $\theta_1$ and $\theta_2$, two plates $\mathcal{P}_0,\mathcal{P}_1$ and we want to approximate $p(\theta_1, \theta_2 | X)$.
 
 A graph template $\mathcal{T}$ can be \textit{grounded} into a HBM $\mathcal{M}$ given some plate cardinalities $\{ \operatorname{Card}(\mathcal{P}_p) \}_{p=0 .. P}$.
 This \textit{grounding} operation instantiates the repeated structures symbolized by the plates: a given RV \textit{template} $\theta_i$ now corresponds to multiple similar \textit{ground} RVs $\{\theta_{i,n}\}_{n=0 .. N_i}$ with the same parametric form, where $N_i = \prod_{\mathcal{P} \in \operatorname{Plates}(\theta_i)} \operatorname{Card}(\mathcal{P})$.
 Template grounding is illustrated in \cref{fig:PAVI_principle}, where $\mathcal{T}$ is instantiated into $\mathcal{M^\text{full}}$.
 We wish to exploit the repeated structure induced by plates.
 
 Given a graph template $\mathcal{T}$, we will instantiate two HBMs. One is our target --or "full"-- model denoted $\mathcal{M^\text{full}}$. This model typically features large plate cardinalities $\operatorname{Card}^\text{full}(\mathcal{P})$, making it computationally intractable. Instead of tackling inference directly for this model, we will instantiate the same template $\mathcal{T}$ into a second HBM $\mathcal{M^\text{redu}}$, the "reduced" model, of tractable plate cardinalities $\operatorname{Card}^\text{redu}(\mathcal{P}) \ll \operatorname{Card}^\text{full}(\mathcal{P})$.
 $\mathcal{M^\text{redu}}$ has the same template as $\mathcal{M^\text{full}}$, meaning the same dependency structure and the same parametric form for its conditional distributions.
 The only difference lies in $\mathcal{M^\text{redu}}$'s reduced cardinalities, resulting in fewer ground RVs, as visible in \cref{fig:PAVI_principle}.
 Our goal is to train over the tractable reduced model $\mathcal{M^\text{redu}}$ to obtain a variational distribution $q$ usable to perform inference over the intractable target model $\mathcal{M^\text{full}}$.
 
 \subsection{Plate amortization}
 \label{sec:plate_amortization}
 In this section we introduce the notion of plate amortization: sharing the parameterization of a conditional density estimator across a model's plates to reduce the parameterization of inference.
 Traditional VI aims at searching for a parametric distribution $q(\Theta ; \phi)$ that will best approximate the posterior distribution of $\Theta$ given a value $\tX_0$ for the $X$: $q(\Theta ; \phi_0) \simeq p(\Theta | X=\tX_0)$ where $\phi_0$ are the optimal weights corresponding to $\tX_0$.
 When presented with a new value $\tX_1$ for X, optimization has to be performed again to search for the weights $\phi_1$, such that $q(\Theta ; \phi_1) \simeq p(\Theta | X=\tX_1)$.
 \textit{Sample amortized} inference \citep{zhang_advances_2019, amortization_gap} aims instead at performing inference in the general case, regressing the weights $\phi$ using an \textit{encoder} $f$ of the observed data $\tX$: $q(\Theta ; \phi=f(\tX)) \simeq p(\Theta | X=\tX)$.
 The cost of learning the weights of the encoder is \textit{amortized} since inference can be performed for any new sample $\tX$ with no additional optimization.
 We propose to exploit the concept of amortization, but to apply it at a different granularity, leading to our notion of \textit{plate amortization}.
 
 Instead of amortizing across the different samples $\tX$ of the observed RV template $X$ we will perform inference amortization across the different ground RVs $\{\theta_{i,n}\}_{n=0 .. N_i}$ corresponding to the same RV template $\theta_i$. 
 Specifically, to a RV template $\theta_i$, we will associate a conditional density estimator $q_{i,\bullet}(\theta_{i, \bullet} ; \phi_i, \bullet)$ with weights $\phi_i$ shared across all the ground RVs $\{\theta_{i,n}\}_{n=0 .. N_i}$.
 The variational posterior for a given ground RV $\theta_{i,n}$ will be an instance of this conditional density estimator, conditioned by an encoding $\tE_{i,n}$: $q_{i,n}(\theta_{i,n} ; \phi_i, \tE_{i,n})$.
 
 The resulting distributions $q_{i,n}$ thus have 2 sets of weights, $\phi_i$ and $\tE_{i,n}$, creating a parameterization trade-off.
 Concentrating all of $q_{i,n}$'s parameterization into $\phi_i$ results in all the ground RVs $\theta_{i,n}$ having almost the same posterior distribution.
 On the contrary, concentrating all of $q_{i,n}$'s parameterization into $\tE_{i,n}$ allows the $\theta_{i,n}$ to have completely different posterior distributions.
 But in a large cardinality setting, this freedom can result in a massive number of weights, proportional to the number of ground RVs times the encoding size.
 This double parameterization is therefore efficient when the majority of the weights for the density estimator $q_{i,n}$ is concentrated into $\phi_i$.
 For instance, casting $q_{i,n}$ as a conditional normalizing flow \citep{papamakarios_normalizing_2019}, the burden of approximating the correct parametric form for the posterior is placed onto $\phi_i$, while $\tE_{i,n}$ can be a lightweight vector of summary statistics specific to each ground RV $\theta_{i,n}$.
 In \cref{sec:training}, we will also see that this shared parameterization has synergies with stochastic training.
 
 \subsection{Variational family design}
 \label{sec:variational_family}
 
 To define our variational family, we will push forward the prior $p(\Theta)$ into the variational distribution $q(\Theta)$.
 This push-forward will be implemented using conditional normalizing flows defined at the graph template level, conditioned by encodings defined at the ground HBM level.
 Consider a RV template $\theta_i$, corresponding to the ground RVs $\theta_{i,n}$.
 In the full HBM $\mathcal{M}^{\text{full}}$, the plate structure indicates that $\theta_i$ is associated to a unique conditional distribution $p_i$ shared across all ground RVs:
 \begin{equation}
     \begin{aligned}
      \log p^{\text{full}}(\Theta, X) &= \sum_{n=0}^{N^{\text{full}}_X} \log p_X(x_n | \pi(x_n)) + \sum_{i=1}^I \sum_{n=0}^{N_i^{\text{full}}} \log p_i(\theta_{i,n} | \pi(\theta_{i,n})) \enspace ,
     \end{aligned}
 \end{equation}
where $\pi(\theta_i^n)$ are the parents of the RV $\theta_i^n$, whose value condition $\theta_i^n$'s distribution.
We indicate with a $\bullet_X$ index all variables related to the observed RVs $X$.
The number of ground RVs $N_i^{\text{full}}$ is the product of the plate cardinalities $\{ \operatorname{Card}^\text{full}(\mathcal{P}) \}_{\mathcal{P} \in \operatorname{Plates}(\theta_i)}$.
To every parameter RV template $\theta_i$, we associate a conditional normalizing flow $\mathcal{F}_i$, parameterized by the weights $\phi_i$.
Every ground RV $\theta_{i,n}$ is in turn associated to a separate encoding $\tE_{i,n}$.
In \cref{fig:PAVI_training}, $\theta_1$ is associated to the flow $\mathcal{F}_1$ pushing forward 2 different ground RVs.
This results in the variational distribution:
\begin{equation}
\label{eq:q_full}
    \begin{aligned}
    \log q^{\text{full}}(\Theta) &= \sum_{i=1}^I \sum_{n=0}^{N_i^{\text{full}}} \log q_{i,n}(\theta_{i,n} | \pi(\theta_{i,n})) \enspace, \\
    \log q_{i,n}(\theta_{i,n} | \pi(\theta_{i,n})) &= -\log \left| \det J_{\mathcal{F}_i}(u_{i,n} ; \phi_i, \tE_{i,n}) \right| + \log p_i(u_{i,n} | \pi(\theta_{i,n})) \enspace, \\
    u_{i,n} &= \mathcal{F}_i^{-1}(\theta_{i,n} ; \phi_i, \tE_{i,n}) \enspace,
    \end{aligned}
\end{equation}
where the distribution $q_{i,n}$ is the push-forward of the prior distribution $p_i$ through the conditional normalizing flow $\mathcal{F}_i$, conditioned by the encoding $\tE_{i,n}$.
This push-forward is illustrated in \cref{fig:PAVI_training}, where flows $\mathcal{F}$ push the RVs $u$ into the RVs $\theta$.
This "cascading" scheme was first introduced by \citet{CF}, and makes $q^{\text{full}}$ inherit the conditional dependencies of the prior $p$.

\subsection{Encoding schemes}
\label{sec:encoding_schemes}
The distributions $q_{i,n}(\theta_{i,n} | \pi(\theta_{i,n}); \phi_i, \tE_{i,n})$ with different ground index $n$ only vary through the value of the encodings $\tE_{i,n}$. We detail two different schemes to derive those encodings:

\textbf{Free plate encodings (PAVI-F)}
In this scheme, $\tE_{i,n}$ are free weights.
We define encodings arrays with the cardinality of the full model $\mathcal{M}^{\text{full}}$, one array $\tE_i = [ \tE_{i,n}]_{n=0..N^{\text{full}}}$ per RV template $\theta_i$.
Using this scheme, the encoding values have the most flexibility, but as a result the variational family's parameterization scales linearly with the cardinalities $\operatorname{Card}^\text{full}(\mathcal{P})$.
Indeed, an additional ground RV in an existing plate necessitates an additional encoding vector.
The resulting weights increment is nevertheless far lighter than the addition of a fully parameterized normalizing flow, as would be the case in the non-plate-amortized regime.
The PAVI-F scheme cannot be sample amortized: when presented with an unseen sample $\tX$, though the value of the weights $\phi_i$ could be kept as an efficient warm start, the optimal value for the encodings $\tE_{i,n}$ would have to be searched again.

\textbf{Deep set encoder (PAVI-E)}
In this scheme the encodings are no longer free weights but obtained processing the observed data $X$ through an encoder $f$: $\tE = f(\tX; \eta)$.
As encoder $f$ we use a \textit{deep-set} architecture exploiting the data's plate-induced permutation invariance --detailed in our supplemental material \citep{deep_sets, ST}.
Encodings $\tE_{i,n}$ no longer are weights for the variational family, and are replaced by the encoder's weights $\eta$.
This scheme furthermore allows for \textit{sample amortization} across different data samples $\tX_0, \tX_1,...$ --see \cref{sec:plate_amortization}.
Note that an encoder will be used to generate the encodings whether the inference is sample amortized or not.

We have defined the architecture $q^{\text{full}}$ to perform inference over the target model $\mathcal{M}^\text{full}$.
Due to the large plate cardinalities $\operatorname{Card}^\text{full}(\mathcal{P})$, it is however computationally impossible to optimize directly over the distribution $q^{\text{full}}$.
In the next section we present a stochastic scheme to overcome this computational hurdle.
\section{PAVI stochastic training}
\label{sec:training}
\subsection{Reduced distribution and loss}
\label{sec:stochastic_training}
\begin{wrapfigure}{}{0.4\textwidth}
    \vspace{-15pt}
    \begin{center}
        \includegraphics[width=0.4\textwidth]{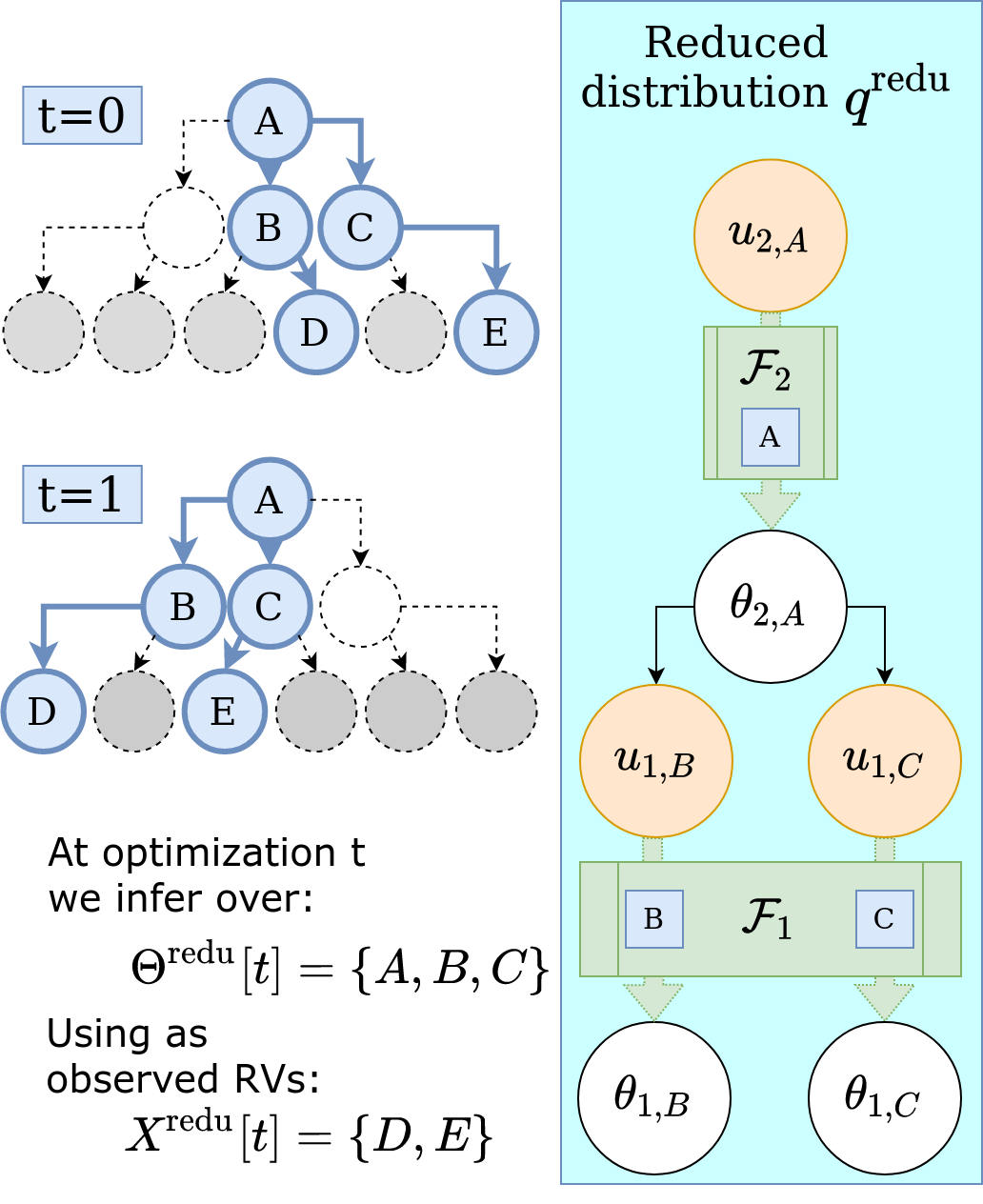}
    \end{center}
    \caption{\textbf{PAVI stochastic training scheme}
    The reduced distribution $q^{\text{redu}}$ features 2 conditional normalizing flows $\mathcal{F}_1$ and $\mathcal{F}_2$ respectively associated to the RV templates $\theta_1$ and $\theta_2$.
    During the stochastic training, $q^{\text{redu}}$ is instantiated over different branchings of the full model $\mathcal{M^\text{full}}$ --highlighted in blue on the left.
    The branchings have the cardinalities of $\mathcal{M^\text{redu}}$ and change at each stochastic training step $t=0,1$.
    The branching determine the encodings $\tE$ conditioning the flows $\mathcal{F}$ --as symbolised by the letters A, B, C-- and the observed data slice --as symbolised by the letters D, E.}
    \label{fig:PAVI_training}
    \vspace{-45pt}
\end{wrapfigure}

Instead of optimizing over the computationally intractable distribution $q^{\text{full}}$, we will use a distribution that has the cardinalities of the reduced model $\operatorname{Card}^\text{redu}(\mathcal{P})$.
At each optimization step $t$, we will randomly select inside $\mathcal{M}^{\text{full}}$ paths of reduced cardinality, as visible in \cref{fig:PAVI_training}.
Selecting paths is equivalent to selecting from $X$ a RV subset of size $N_X^\text{redu}$, denoted $X^{\text{redu}}[t]$.
We subsequently select from $\Theta$ the RV set $\Theta^{\text{redu}}[t]$ of ascendants and descendants of $X^{\text{redu}}[t]$.
For a given $\theta_i$, we denote as $\mathcal{B}^{\text{redu}}_{i}[t]$ the resulting batch of selected ground RVs, of size $N_i^{\text{redu}}$.
Inferring over $\Theta^{\text{redu}}[t]$, we will simulate the fact that we train over the distribution $q^{\text{full}}$, resulting in the distribution:
\begin{equation}
\label{eq:q_stoc}
    \begin{aligned}
     \log q^{\text{redu}}(\Theta^{\text{redu}}[t]) &= \sum_{i=1}^I \frac{N_i^{\text{full}}}{N_i^{\text{redu}}} \sum_{\mathclap{\qquad n \in \mathcal{B}^{\text{redu}}_{i}[t]}} \log q_{i,n}(\theta_{i,n} | \pi(\theta_{i,n}))
    \end{aligned}
\end{equation}
where the factor $N_i^{\text{full}} / N_i^{\text{redu}}$ simulates that we observe as many ground RVs as in the full HBM $\mathcal{M}^{\text{full}}$ by repeating the ground RVs from $\mathcal{M}^{\text{redu}}$ \citep{SVI}.
Similarly, the loss used at the optimization step $t$ is the reduced ELBO constructed using $X^{\text{redu}}[t]$ as observed RVs:
\begin{equation}
\label{eq:ELBO_stoc}
    \begin{aligned}
    \log p^{\text{redu}}(X^{\text{redu}}[t], \Theta^{\text{redu}}[t]) &= \frac{ N_X^{\text{full}}}{N_X^{\text{redu}}} \sum_{\mathclap{\qquad n \in \mathcal{B}^{\text{redu}}_{X}[t]}} \log p_X(x_n | \pi(x_n)) + \sum_{i=1}^I \frac{N_i^{\text{full}}}{N_i^{\text{redu}}} \sum_{\mathclap{\qquad n \in \mathcal{B}^{\text{redu}}_{i}[t]}} \log p_i(\theta_{i,n} | \pi(\theta_{i,n})) \\
     \operatorname{ELBO}^{\text{redu}}[t] &= \mathbb{E}_{\Theta^{\text{redu}} \sim q^{\text{redu}}}\left[ \log p^{\text{redu}}(X^{\text{redu}}[t], \Theta^{\text{redu}}[t])  - \log q^{\text{redu}}(\Theta^{\text{redu}}[t]) \right]
    \end{aligned}
\end{equation}
This scheme can be viewed as the instantiation of $\mathcal{M}^\text{redu}$ over batches of $\mathcal{M}^\text{redu}$'s ground RVs.
In \cref{fig:PAVI_training} we can see that $q^\text{redu}$ has the cardinalities of $\mathcal{M}^\text{redu}$, and replicates its conditional dependencies.
The resulting training is analogous to the usage of stochastic VI \citep{SVI} over $\mathcal{M}^{\text{full}}$, generalized with multiple hierarchies and using minibatches of ground RVs.
Our novelty lies in the interaction of this stochastic scheme with plate amortization, as explained in the next section.

\subsection{Sharing learning across plates}
\label{sec:sharing_learning}
In a traditional stochastic VI training, every ground RV $\theta_{i,n}$ corresponding to the same template $\theta_i$ is associated to individual weights.
Those weights are trained only when $\theta_{i,n}$ is visited by the algorithm, that is to say at an optimization step $t$ when $n \in \mathcal{B}^{\text{redu}}_{i}[t]$.
In the context of very large model plates, this event can become rare.
If $\theta_{i,n}$ is furthermore associated to a highly-parameterized density estimator --such as a normalizing flow-- many optimization steps can be required for the distribution $q_{i,n}$ to converge.
The combination of those two items can lead to a slow training.

Instead, our idea is to share the learning across the ground RVs $\theta_{i,n}$.
Indeed, due to the problem's plate structure, we consider the inference over those ground RVs as different instances of a common density estimation task.
The precise implementation of this shared learning depends on the chosen encoding scheme --as described in \cref{sec:encoding_schemes}:

\textbf{Conditional flow weight sharing (PAVI-F)}
As seen in \cref{sec:variational_family}, a large part of the parameterization of the density estimators $q_{i,n}(\theta_{i,n} | \pi(\theta_{i,n}); \phi_i, \tE_{i,n})$ is mutualized via the plate-wide-shared weights $\phi_i$.
At each optimization step $t$, the encodings $\tE_{i,n}$ corresponding to $n \in \mathcal{B}^{\text{redu}}_{i}[t]$ are sliced from larger encoding arrays $\tE_i = [ \tE_{i,n}]_{n=0..N^{\text{full}}}$ and are optimized for along with the weights $\phi_i$.
This means that most of the weights of the flows $\mathcal{F}_i$ --concentrated in $\phi_i$-- are trained at every optimization step, across all the selected batches $\mathcal{B}^{\text{redu}}_{i}[t]$.
This can result in drastically faster convergence, as demonstrated in our experiments.
In \cref{fig:PAVI_training}, at $t=0$, $\mathcal{B}^{\text{redu}}_{1}[0] = \{ 1, 2 \}$ and the trained encodings are therefore $\{ \tE_{1,1}, \tE_{1,2} \}$, and at $t=1$  $\mathcal{B}^{\text{redu}}_{1}[1] = \{ 0, 1 \}$ and the used encodings are $\{ \tE_1^0, \tE_1^1 \}$.
The weights $\phi_1$ and $\phi_2$ of the flows $\mathcal{F}_1$ and $\mathcal{F}_2$ are trained at both steps $t=1$ and $t=2$.
At inference, instead of slicing the encoding arrays, the full arrays $\tE_i$ are used to obtain the distribution $q^{\text{full}}$.

\textbf{Encoder set size generalization (PAVI-E)}
The PAVI-E scheme also benefits from the sharing of the weights $\phi_i$.
In addition, it doesn't cast the encodings $\tE_{i,n}$ as free weights, but as the output of a parametric encoder $f(\bullet;\eta)$.
As a result, at training all the architecture's weights --$\phi_i$ and $\eta$-- are trained at every optimization step $t$.
At inference, to generate the full encoding arrays $\tE_i = [ \tE_{i,n}]_{n=0..N^{\text{full}}}$ to plug into $q^{\text{full}}$, this scheme builds up on a property of the particular deep-set-like architecture we use for the encoder $f$: \textit{set size generalization} \citep{deep_sets}.
Through training, the encoder $f$ learnt a hierarchy of permutation-invariant functions over $\operatorname{Card}^\text{redu}(\mathcal{P})$-sized sets of data points.
At inference, we instead apply the trained encoder to sets of size $\operatorname{Card}^\text{full}(\mathcal{P})$:
\begin{equation}
\label{eq:E-SSG}
    \begin{aligned}
    \textit{At training:} \qquad  \tE_{i,n} &= f_{i,n}(\tX^{\text{redu}}[t]) &\qquad \text{for} \quad n \in \mathcal{B}^\text{redu}_{i}[t] \\
    \textit{At inference:} \qquad \tE_{i,n} &= f_{i,n}(\tX) &\qquad \text{for} \quad n = 0 .. N_i^{\text{full}}
    \end{aligned}
\end{equation}
where $\tX^{\text{redu}}[t]$ denotes the observed data corresponding to $X^{\text{redu}}[t]$.
This property --learning an encoder over small sets to use it over large sets-- is very strong, especially in the sample amortized context. Benefiting from set size generalization, we can effectively train a sample amortized variational family over the lightweight model $\mathcal{M}^{\text{redu}}$, and obtain "for free" a sample amortized variational family for the heavyweight model $\mathcal{M}^{\text{full}}$.

\textbf{Summary}
In \cref{sec:archi} we proposed an architecture sharing its parameterization across a model's plates.
In \cref{sec:training} we proposed a stochastic scheme to train this architecture over batches of data of reduced cardinality.
Across those data batches, we share the learning of density estimators, resulting in the fast training of a variational posterior $q^{\text{full}}$, as demonstrated in the following experiments.
\section{Results and discussion}
\label{sec:exps}
All experiments were performed using the Tensorflow Probability library \citep{dillon_tensorflow_2017}, on computational cluster nodes equipped with a Tesla V100-16Gb GPU and 4 AMD EPYC 7742 64-Core processors. VRAM intensive experiments in \cref{fig:scaling} were performed on an Ampere 100 PCIE-40Gb GPU.
Throughout this section we focus on the usage of the ELBO metric, as a proxy to the KL divergence between the variational posterior and the unknown true posterior.
ELBO is measured across 20 different data samples $\tX$, with 5 random seeds per sample.
The ELBO allows to compare the relative performance of different architectures on a given inference problem.
In our supplemental material we also provide with sanity checks to assess the quality of the obtained results.

\subsection{Plate amortization and convergence speed}
\label{sec:exp_convergence_speed}
In this experiment, we illustrate how plate amortization results in faster convergence.
We consider the following Gaussian Random Effects model (GRE):
\begin{equation}
\label{eq:GRE}
\begin{aligned}
    \substack{\forall n_1=1..\operatorname{Card}(\mathcal{P}_1) \\ \forall n_0=1..\operatorname{Card}(\mathcal{P}_0)} \quad  X_{n_1,n_0} |\theta_{1,n_1} &\sim \mathcal{N}(\theta_{1,n_1}, \sigma_x^2) \\
   \forall n_1=1..\operatorname{Card}(\mathcal{P}_1) \quad  \theta_{1,n_1} | \theta_{2,0} &\sim \mathcal{N}(\theta_{2,0}, \sigma_1^2) \qquad 
    \theta_{2,0} &\sim \mathcal{N}(\vec 0_D, \sigma_2^2) \enspace ,
\end{aligned}
\end{equation}
where $D$ represents the feature size of the data \tX, determining the dimensionality of the group means $\theta_1$ and of the population means $\theta_2$ as D-dimensional Gaussians.
We opted in this equation for a more practical double indexing scheme instead of a simple indexing as in our methods.
The GRE model features two nested plates: the group plate $\mathcal{P}_1$ and the sample plate $\mathcal{P}_0$ as in \cref{fig:PAVI_principle}.
Performing inference over this HBM, the objective is to retrieve the posterior distribution of the group means $\theta_1$ and the population mean $\theta_2$ given the observed sample $X$.

\begin{figure}
    \centering
    \begin{subfigure}[t]{0.48\textwidth}
        \centering
        \includegraphics[width=\textwidth]{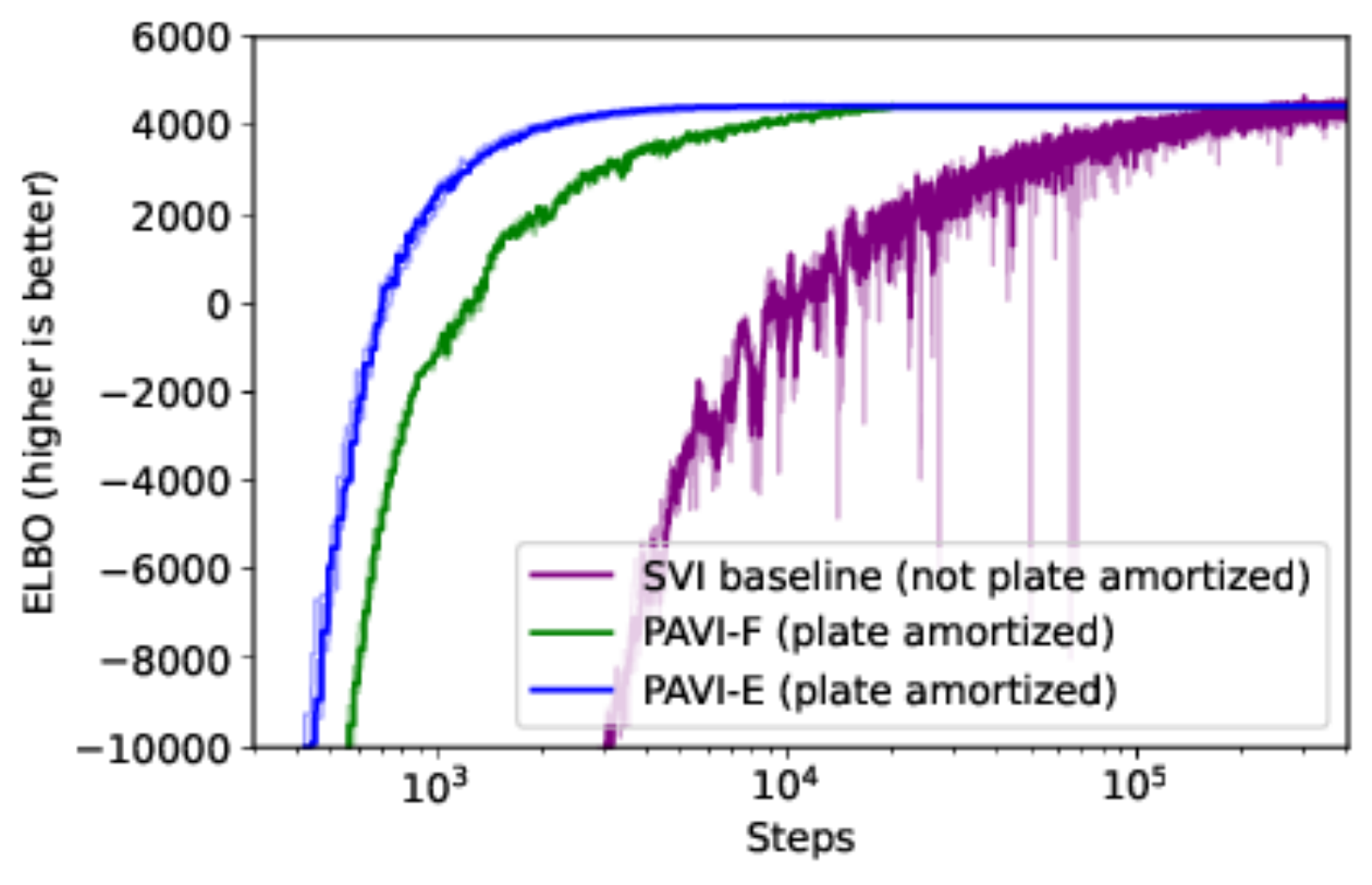}
    \end{subfigure}
    \hfill
    \begin{subfigure}[t]{0.47\textwidth}
        \centering
        \includegraphics[width=\textwidth]{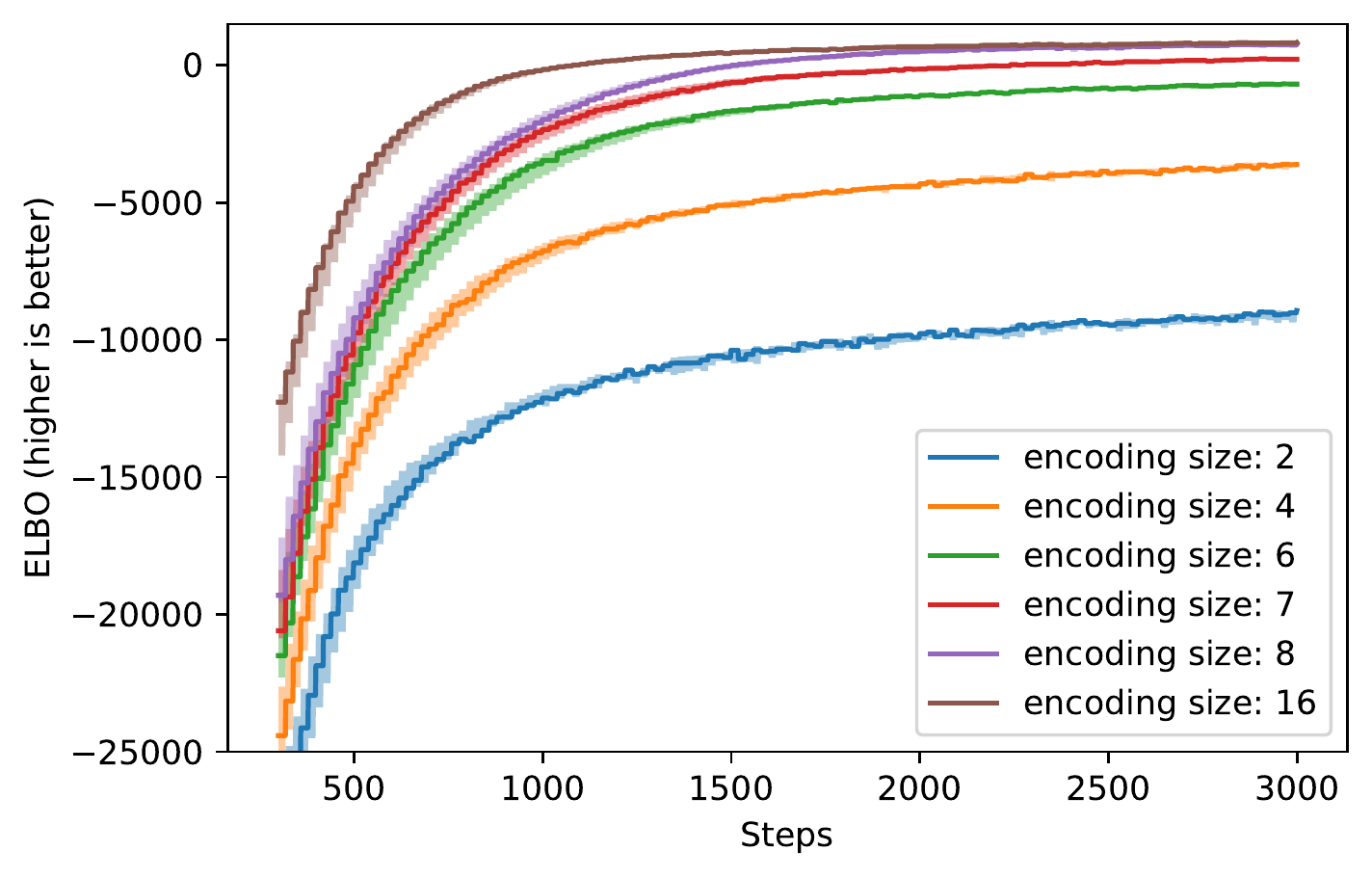}
    \end{subfigure}
    \vspace{-5pt}
    \caption{\textbf{Left panel: Plate amortization increases convergence speed} Plot of the ELBO (higher is better) as a function of the optimization steps (log-scale) for our methods PAVI-F (in green) and PAVI-E (in blue) versus a non-plate-amortized baseline (in purple).
    Due to plate amortization, our method converges orders of magnitude faster to the same asymptotic ELBO as its non-plate-amortized counterpart.;
    \textbf{Right panel: Encodings as ground RVs summary statistics} Plot of the ELBO (higher is better) as a function of the optimization steps for the PAVI-F architecture with increasing encoding sizes.
    As the encoding size augments, so does the asymptotic performance, until reaching the dimensionality of the posterior's sufficient statistics ($D=8$), after which performance plateaus.
    Encoding size allows for a clear trade-off between memory footprint and inference quality.}
    \label{fig:convergence_speed}
    \label{fig:encoding_size}
    \vspace{-15pt}
\end{figure}

Here we set $D=8$, $\operatorname{Card}^{\text{full}}(\mathcal{P}_1)=100$ and $\operatorname{Card}^{\text{redu}}(\mathcal{P}_1)=2$.
We compare our PAVI architecture to a stochastic non-plate-amortized baseline with the same architecture as PAVI \citep{SVI}.
The main difference is that every ground RV $\theta_i^n$ is associated in the baseline to an individual flow $\mathcal{F}_{i,n}$ instead of sharing the same flow $\mathcal{F}_i$ --as described in \cref{sec:variational_family}.
\Cref{fig:convergence_speed} (left) displays the evolution of the ELBO for the baseline and PAVI with free encoding (PAVI-F) and with deep set encoders (PAVI-E).
We see that for both plate amortized methods, the convergence speed to an asymptotic ELBO equals to the one of the non-plate-amortized baseline is orders of magnitudes faster, and numerically more stable.
This stems from the individual flows $\mathcal{F}_{i,n}$ only being trained when the corresponding $\theta_{i,n}$ is visited by the stochastic training, while the shared flow $\mathcal{F}_i$ is updated at every optimization step in PAVI.
We also note that the PAVI-E scheme has a faster convergence than the PAVI-F scheme, sharing not only the training of the conditional flows, but also of the encoder through the stochastic optimization steps.
In practice however, the additional compute implied by the encoder results step of longer duration, and ultimately in slower convergence, as illustrated in \cref{sec:exp_scaling}.

\subsection{Impact of encoding size}
\label{sec:exp_encoding_size}
Now we illustrate the role of encodings as ground RV's posterior summary statistics --as described in \cref{sec:plate_amortization}.
We use the GRE HBM detailed in \cref{eq:GRE}, using $D=8$, $\operatorname{Card}^{\text{full}}(\mathcal{P}_1)=20$ and $\operatorname{Card}^{\text{redu}}(\mathcal{P}_1)=2$.
We use a single PAVI-F architecture, varying the dimensionality of the encodings $\tE_{i,n}$ --see \cref{sec:variational_family}. Due to plate amortization, this encoding size determines how much individual information each ground RV $\theta_{i,n}$ is associated to. The size of the encodings --varying from $2$ to $16$-- is to be compared with the dimensionality of the problem, in this case $D=8$. Indeed, in the GRE context, $D=8$ corresponds to the dimensionality of the sufficient statistics needed to reconstruct the posterior distribution of a given group mean --all other statistics such as the posterior variance being shared between all the group means.
\Cref{fig:encoding_size} (right) shows how the asymptotic performance steadily increases when the encoding size augments, before plateauing once reaching the sufficient summary statistic size $D=8$.
Interestingly, increasing the encoding size also leads to faster convergence: redundancy in the encoding can likely be exploited in the optimization.
Encoding size appears as a straightforward hyperparameter allowing to trade inference quality for computational efficiency.
It is also interesting to notice that increasing the encoding size leads experimentally to diminishing returns in terms of performance. This property can be exploited in large dimensionality settings to drastically reduce the memory footprint of inference while maintaining acceptable performance.

\subsection{Scaling with plate cardinalities}
\begin{figure}[t]
    \centering
    \includegraphics[width=0.95\textwidth]{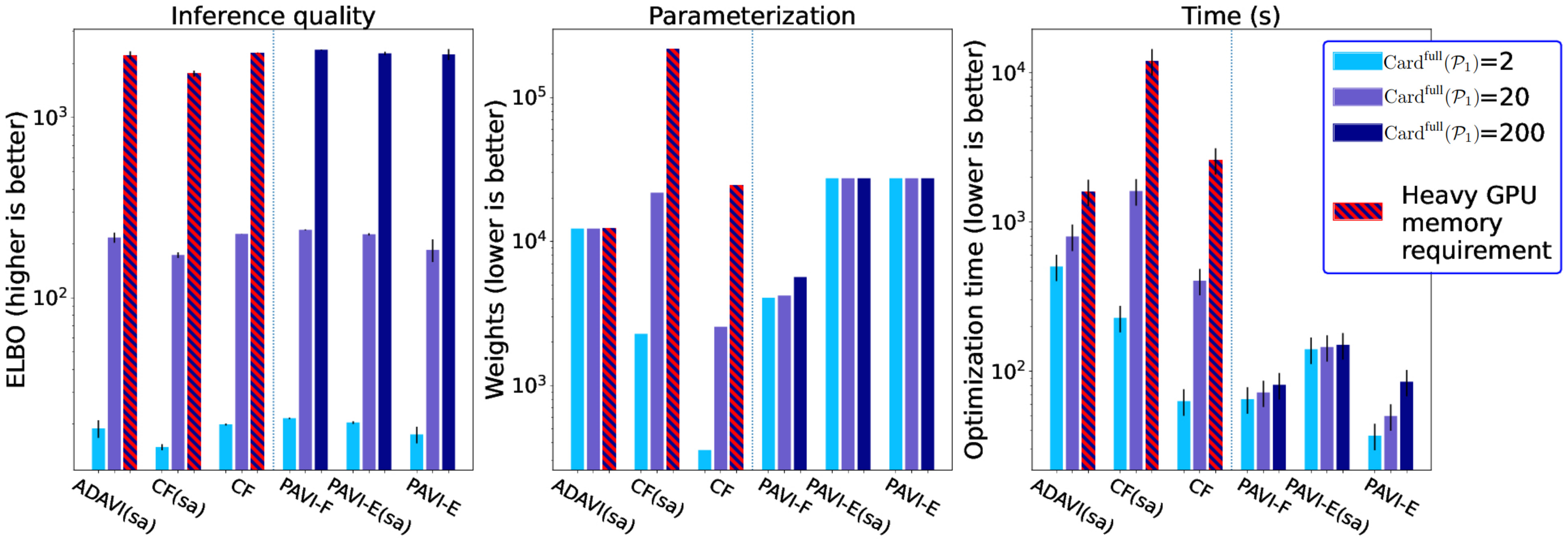}
    \caption{\textbf{PAVI provides with favorable parameterization and training time as the cardinality of the target model augments}
    Our architecture PAVI is displayed on the right of each panel.
    We augment the cardinality $\operatorname{Card}^{\text{full}}(\mathcal{P}_1)$ of the GRE model --described in \cref{eq:GRE}.
    While doing so, we compare 3 different metrics:
    \textit{In the first panel:} inference quality, as measured by the ELBO.
    None of the presented SOTA architecture's performance degrades as the cardinality of the problem augments.
    \textit{In the second pannel:} parameterization, comparing the number of trainable weights of each architecture.
    PAVI --similar to ADAVI-- displays a constant number of weights as the cardinality of the problem increases --or almost constant for PAVI-F.
    \textit{Third panel:} GPU training time.
    Benefiting from learning across plates, PAVI has a short and almost constant training time as the cardinality of the problem augments.
    At $\operatorname{Card}^{\text{full}}(\mathcal{P}_1)=200$, CF and ADAVI required large GPU memory, a constraint absent from PAVI due to its stochastic training.}
    \label{fig:scaling}
    \vspace{-15pt}
\end{figure}
\label{sec:exp_scaling}
Now we put in perspective the gains from plate amortization when scaling up an inference problem's cardinality.
We consider the GRE model in \cref{eq:GRE} with $D=2$ and augment the plate cardinalities $(\operatorname{Card}^{\text{full}}(\mathcal{P}_1),\operatorname{Card}^{\text{redu}}(\mathcal{P}_1)):\ (2, 1) \rightarrow (20, 5) \rightarrow (200, 20)$.
In doing so, we augment the number of estimated parameters $\Theta:\ 6 \rightarrow 42 \rightarrow 402$.

\textbf{Baselines} We compare our PAVI architecture against 2 state-of-the-art baselines:
\textit{Cascading Flows} (CF) \citep{CF} is a non-plate-amortized structured VI architecture improving on the baseline presented in \cref{sec:exp_convergence_speed};
ADAVI \citep{ADAVI} is a structured VI architecture with constant parameterization with respect to a problem's cardinality, but large training times and memory footprint.
For all architectures, we indicate with the suffix \textit{(sa) sample amortization}, corresponding to the classical meaning of amortization, as detailed in \cref{sec:plate_amortization}.
More details can be found in our supplemental material.

As the cardinality of the problem augments, \cref{fig:scaling} shows how PAVI maintains a state-of-the-art inference quality, while being more computationally attractive.
Specifically, in terms of \textit{parameterization}, both ADAVI and PAVI-E provide with a heavyweight but constant parameterization as the cardinality $\operatorname{Card}^{\text{full}}(\mathcal{P}_1)$ of the problem augments.
Comparatively, both CF and PAVI-F's parameterization scale linearly with $\operatorname{Card}^{\text{full}}(\mathcal{P}_1)$, but with a drastically lighter augmentation for PAVI-F.
Indeed, for an additional ground RV, CF requires an additional fully parameterized normalizing flow, whereas PAVI-F only requires an additional lightweight encoding vector.
In detail, PAVI-F's parameterization due to the plate-wide-shared $\phi_1$ represents a constant $\approx 2k$ weights, while the part due to the encodings $\tE_{1,n}$ grows linearly from $16$ to $160$ to $1.6k$ weights.
Note that PAVI's stochastic training also allows for a controlled GPU memory during optimization, removing the need for a larger memory as the cardinality of the problem augments --a hardware constraint that can become unaffordable at very large cardinalities.
In terms of \textit{convergence speed}, PAVI benefits from plate amortization to have orders of magnitude faster convergence.
Plate amortization is particularly significant for the PAVI-E(sa) scheme, in which a sample-amortized variational family is trained over a dataset of reduced cardinality, yet performs "for free" inference over a HBM of  large cardinality.
Maintaining $\operatorname{Card}^{\text{redu}}(\mathcal{P}_1)$ constant while $\operatorname{Card}^{\text{full}}(\mathcal{P}_1)$ augments allows for a constant parameterization \textit{and training time} as the cardinality of the problem augments.
The effect of plate amortization is particularly noticeable at $\operatorname{Card}^{\text{full}}(\mathcal{P}_1)=200$ between the PAVI(sa) and CF(sa) architectures, where PAVI performs amortized inference with $10\times$ fewer weights and $100\times$ lower training time.
Scaling even higher the cardinality of the problem --$\operatorname{Card}^{\text{full}}(\mathcal{P}_1)=2000$ for instance-- renders ADAVI and CF computationally intractable to use, while PAVI maintains a light memory footprint, and a short training time, as exemplified in the next experiment.

\subsection{Application: fMRI -- parcellation of Broca's area over a large subject cohort}
\label{sec:exp_fMRI}
\begin{figure}
    \centering
    \includegraphics[width=0.95\textwidth]{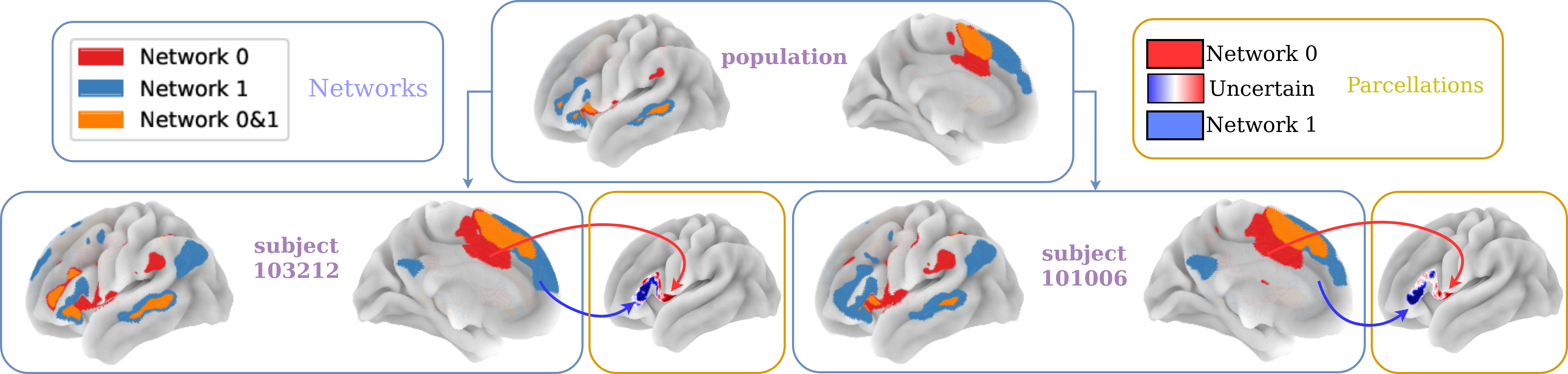}
    \caption{\textbf{Probabilistic parcellation of Brocas's area}
    PAVI can be applied in the challenging context of Neuroimaging population studies.
    For a cohort of $1000$ subjects, 2 of which are represented here --in the bottom 2 items-- we present 2 results.
    First, \textit{connectivity networks} with the brain's left hemisphere --left purple items: this represents the zones of the brain to which the vertices with each label are "wired" to.
    Second, Broca's area probabilistic \textit{parcellation} --rightmost orange items: we cluster the brain's vertices, associating them to a given \textit{connectivity network}.
    Our Bayesian method features a notion of uncertainty: coloring transitions from red to an uncertain white to blue, representing the probability of a given vertex to belong to one connectivity network or the other.}
    \label{fig:broca}
    \vspace{-15pt}
\end{figure}

To illustrate the usefulness of our method, we apply PAVI to a challenging Neuroimaging example: a population study for Broca's area's functional \textit{parcellation}.
A \textit{parcellation} of a brain region aims at clustering brain vertices into different \textit{connectivity networks}: labels describing the vertices' co-activation with the rest of the brain --as measured using functional Magnetic Resonance Imaging (fMRI).
Different subjects can exhibit a strong variability, as visible in \cref{fig:broca}.
However, fMRI has a costly acquisition --meaning that few noisy data is usually gathered for a given subject.
It is thus essential to combine the information from different subjects and to have a notion of uncertainty in the obtained results.
Those 2 points motivate our usage of Hierarchical Bayesian Models and VI in the Neuroimaging context \citep{kong_spatial_2018}: we wish to obtain the posterior distribution of connectivity networks and vertex labels, combining fMRI measurement over a large cohort of subjects.
In practice, we use the HCP dataset \citep{HCP}: $2$ acquisition sessions for a cohort of $1000$ subjects, with thousands of measurements per subject, for a total parameter space $\Theta$ of over a million parameters.
We use a model with 3 plates: subjects, measurement sessions and brain vertices.
In this high plate cardinality regime, none of the state-of-the-art baselines presented in \cref{sec:exp_scaling} --CF, ADAVI-- can computationally tackle inference.
In terms of convergence speed, despite the massive dimensionality of the problem, thanks to plate amortization PAVI converges in a dozen epochs, under an hour of GPU time.
The results of our method are visible in \cref{fig:broca}, supporting the hypothesis of a functional bi-partition of Broca's area into a posterior part involved in phonology and an anterior part involved in lexical/semantic processing - following the anatomical partition between \textit{pars opercularis} and \textit{pars triangularis} \citep{heimEffectiveConnectivityLeft2009, zhangConnectingConceptsBrain2020}.

\subsection{\textbf{Conclusion}}
In this work we present the novel PAVI architecture, combining a structured variational family and a stochastic training scheme.
PAVI is based the concept of plate amortization, allowing to share parameterization and learning across a model's plates.
We demonstrated the positive impact of plate amortization on training speed and scaling to large plate cardinality regimes, making a significant step towards scalable, expressive Variational Inference.

\begin{ack}
This work was supported by the ERC-StG NeuroLang ID:757672.
\end{ack}

\bibliographystyle{iclr_style}
\bibliography{references}



\newpage
\appendix

\section*{Supplemental Material}

\setcounter{figure}{0}
\setcounter{equation}{0}
\renewcommand\thefigure{\thesection.\arabic{figure}}
\renewcommand\theequation{\thesection.\arabic{equation}}

\section{Supplemental methods}
\subsection{PAVI implementation details}
\subsubsection{Plate branchings and stochastic training}
\label{sec:plate_branching}
As exposed in \cref{sec:stochastic_training}, at each optimization step $t$ we randomly select branchings inside the full model $\mathcal{M}^\text{full}$, branchings over which we "instantiate" the reduced model $\mathcal{M}^\text{redu}$.
In doing so, we define batches $\mathcal{B}_{i}[t]$ for the RV templates $\theta_i$.
Those batches have to be "coherent" with one another: they have to respect the conditional dependencies of the original model $\mathcal{M}^{\text{full}}$.
To ensure this, during the stochastic training we do not sample RVs directly but plates:
\begin{enumerate}
    \item For every plate $\mathcal{P}_p$, we sample without replacement $\operatorname{Card}^{\text{redu}}(\mathcal{P}_p)$ indices amongst the $\operatorname{Card}^{\text{full}}(\mathcal{P}_p)$ possible indices.
    \item Then, for every RV template $\theta_i$, we select the ground RVs $\theta_{i,n}$ corresponding to the sampled indices for the plates $\operatorname{Plates}(\theta_i)$.
    \item The selected ground RVs $\theta_{i,n}$ constitute the set $\Theta^\text{redu}[t]$ of parameters appearing in \cref{eq:q_stoc}. The same procedure yields the RV subset $X^\text{redu}[t]$ and the data slice $\tX^\text{redu}[t]$.
\end{enumerate}
This stochastic strategy also applies to the selected encoding scheme --described in \cref{sec:encoding_schemes}-- as detailed in the next sections.

\subsubsection{PAVI-F details}
\label{sec:PAVI-F_details}
In \cref{sec:encoding_schemes} we refer to encodings $\tE_i = [ \tE_{i,n}]_{n=0..N^{\text{full}}}$ corresponding to RV templates $\theta_i$.
In practice, we have some amount of sharing for those encodings: instead of defining separate encodings for every RV template, we define encodings for every \textit{plate level}.
A plate level is a combination of plates with at least one parameter RV template $\theta_i$ belonging to it:
\begin{equation}
    \begin{aligned}
    \operatorname{PlateLevels} &= \{ (\mathcal{P}_k..\mathcal{P}_l) = \operatorname{Plates}(\theta_i) \}_{\theta_i \in \Theta}
    \end{aligned}
\end{equation}
For every plate level, we construct a large encoding array with the cardinalities of the full model $\mathcal{M}^{\text{full}}$:
\begin{equation}
    \begin{aligned}
    \operatorname{Encodings} &= \{ (\mathcal{P}_k..\mathcal{P}_l) \mapsto \mathbb{R}^{\operatorname{Card}^{\text{full}}(\mathcal{P}_k) \times .. \times \operatorname{Card}^{\text{full}}(\mathcal{P}_l) \times D} \}_{(\mathcal{P}_k..\mathcal{P}_l) \in \operatorname{PlateLevels}} \\
    \tE_i &= \operatorname{Encodings}(\operatorname{Plates}(\theta_i))
    \end{aligned}
\end{equation}
Where $D$ is an encoding size that we kept constant to de-clutter the notation but can vary between plate levels.
The encodings for a given ground RV $\theta_{i,n}$ then correspond to an element from the encoding array $\tE_i$. 

\subsubsection{PAVI-E details}
\label{sec:PAVI-E_details}
In the PAVI-E scheme, encodings are not free weights but the output of en encoder $f(\bullet, \eta)$ applied to the observed data $\tX$.
In this section we detail the design of this encoder.

As in the previous section, the role of the encoder will be to produce one encoding per plate level.
We start from a dependency structure for the plate levels:
\begin{equation}
    \begin{aligned}
    \forall (\mathcal{P}_a..\mathcal{P}_b) &\in \operatorname{PlateLevels} \enspace, \\
    \forall (\mathcal{P}_c..\mathcal{P}_d) &\in \operatorname{PlateLevels} \enspace, \\
    (\mathcal{P}_a..\mathcal{P}_b) \in \pi((\mathcal{P}_c..\mathcal{P}_d)) &\Leftrightarrow \substack{\exists \theta_i/\operatorname{Plates}(\theta_i) = (\mathcal{P}_a..\mathcal{P}_b) \\ \exists \theta_j/\operatorname{Plates}(\theta_j) = (\mathcal{P}_c..\mathcal{P}_d)} / \theta_j \in \pi(\theta_i)
    \end{aligned}
\end{equation}
note that this dependency structure is in the "backward" direction: a plate level will be the parent of another plate level, if the former contains a RV who has a child in the latter.
We therefore obtain a plate level dependency structure that "reverts" the conditional dependency structure of the graph template $\mathcal{T}$.
To avoid redundant paths in this dependency structure, we take the maximum branching of the obtained graph.

Given the plate level dependency structure, we will recursively construct the encodings, starting from the observed data:
\begin{equation}
    \begin{aligned}
    \forall x \in X:\quad \operatorname{Encodings}(\operatorname{Plates}(x)) = \rho (\textbf{x})
    \end{aligned}
\end{equation}
where $\textbf{x}$ is the observed data for the RV $x$, and $\rho$ is a simple encoder that processes every observed ground RV's value independently through an identical multi-layer perceptron.
Then, until we have exhausted all plate levels, we process existing encodings to produce new encodings:
\begin{equation}
    \begin{aligned}
    \forall (\mathcal{P}_k..\mathcal{P}_l) &\in \operatorname{PlateLevels} / \not\exists x \in X, \operatorname{Plates}(x) = (\mathcal{P}_k..\mathcal{P}_l): \\
    \operatorname{Encodings}((\mathcal{P}_k..\mathcal{P}_l)) &= g(\operatorname{Encodings}(\pi(\mathcal{P}_k..\mathcal{P}_l)))
    \end{aligned}
\end{equation}
where $g$ is the composition of attention-based deep-set networks called \textit{Set Transformers} \citep{ST, deep_sets}.
For every plate $\mathcal{P}_p$ present in the parent plate level but absent in the child plate level, $g$ will compute summary statistics \textit{across} that plate, effectively contracting the corresponding batch dimensionality in the parent encoding \citep{ADAVI}.

In the case of multiple observed RVs, we run this "backward pass" independently for each observed data --with one encoder per observed RV.
We then concatenate the resulting encodings corresponding to the same plate level.

For more precise implementation details, we invite the reader to consult the codebase released with this supplemental material.

\subsection{PAVI algorithms}

More technical details can be found in the codebase provided with this supplemental material.

\subsubsection{Architecture build}
\begin{algorithm}[H]
\caption{PAVI architecture build}
\label{alg:arch}
\KwIn{Graph template $\mathcal{T}$, plate cardinalities $\{ (\operatorname{Card}^{\text{full}}(\mathcal{P}_p), \operatorname{Card}^{\text{redu}}(\mathcal{P}_p)) \}_{p=0..P}$, encoding scheme}
\KwOut{$q^\text{full}$ distribution}
\For{$i=1..I$}{
    Construct conditional flow $\mathcal{F}_i$\;
    Define conditional posterior distributions $q_{i,n}$ as the push-forward of the prior via $\mathcal{F}_i$, following \cref{eq:q_full}\;
}
Combine the $q_{i,n}$ distributions following the cascading flows scheme, as in \cref{sec:variational_family} \citep{CF} \;
\uIf{PAVI-F encoding scheme}{
    Construct encoding arrays $\{  \tE_i = \left[ \tE_{i,n} \right]_{n=0..N_i^{\text{full}}} \}_{i=1..I}$ as in \cref{sec:PAVI-F_details} \;
}
\ElseIf{PAVI-E encoding scheme}{
    Construct encoder $f$ as in \cref{sec:PAVI-E_details} \;
}
\end{algorithm}

\subsubsection{Stochastic training}
\begin{algorithm}[H]
\caption{PAVI stochastic training}
\label{alg:train}
\KwIn{Untrained architecture $q^\text{full}$, observed data $\tX$, encoding scheme, number of steps $T$}
\KwOut{trained architecture $q^\text{full}$}
\For{$t=0..T$}{
    Sample plate indices to define the batches $\mathcal{B}_i[t]$, the latent $\Theta^\text{redu}[t]$ and the observed $X^\text{redu}[t]$ and $\tX^\text{redu}[t]$, following \cref{sec:plate_branching} \;
    Define reduced distribution $p^\text{redu}$ \;
    \uIf{PAVI-F encoding scheme}{
        Collect encodings $\tE_{i,n}$ by slicing from the arrays $\tE_i$ the elements corresponding to the batches $\mathcal{B}_i[t]$ \;
    }
    \ElseIf{PAVI-E encoding scheme}{
        Compute encodings as $\tE = f(\tX^\text{redu}[t]) $\;
    }
    Feed obtained encodings into $q^\text{redu}$ \;
    Compute reduced ELBO as in \cref{eq:ELBO_stoc}, back-propagate its gradient \;
    Update conditional flow weights $\{ \phi_i \}_{i=1..I}$\;
    \uIf{PAVI-F encoding scheme}{
        Update encodings $\{ \tE_{i,n} \}_{i=1..I, n \in \mathcal{B}_{i, t}}$\;
    }
    \ElseIf{PAVI-E encoding scheme}{
        Update encoder weights $\eta$\;
    }
}
\end{algorithm}

\subsubsection{Inference}
\begin{algorithm}[H]
\caption{PAVI inference}
\label{alg:infer}
\KwIn{trained architecture $q^\text{full}$, observed data $\tX$, encoding scheme}
\KwOut{approximate posterior distribution}
\uIf{PAVI-F encoding scheme}{
    Collect full encoding arrays $\tE_i$ \;
}
\ElseIf{PAVI-E encoding scheme}{
    Compute encodings as $\tE = f(\tX)$ using set size generalization \;
}
Feed obtained encodings into $q^\text{full}$ \;
\end{algorithm}

\subsection{Inference gaps}
In terms of inference quality, the impact of our architecture can be formalized following the \textit{gaps} terminology \citep{amortization_gap}.
Consider a joint distribution $p(\Theta, X)$, and a value $\tX$ for the RV template $X$.
We pick a variational family $\mathcal{Q}$, and in this family look for the parametric distribution $q(\Theta; \phi)$ that best approximates $p(\Theta | X=\tX)$.
Specifically, we want to minimize the Kulback-Leibler divergence \citep{blei_variational_2017} between our variational posterior and the true posterior, that \citet{amortization_gap} refer to as the \textit{gap} $\mathcal{G}$:
\begin{equation}
\begin{aligned}
    \mathcal{G} &= \operatorname{KL}(q(\Theta;\phi) || p(\Theta | X)) \\
            &= \log p(X) - \operatorname{ELBO}(q; \phi)
\end{aligned}
\end{equation}
We denote $q^{*}(\Theta; \phi^*)$ the optimal distribution inside $\mathcal{Q}$ that minimizes the KL divergence with the true posterior:
\begin{equation}
    \begin{aligned}
    \mathcal{G}_{\text{approx}}(\mathcal{Q}; \phi^*) &= \log p(X) - \operatorname{ELBO}(q^*; \phi^*) \\
    &\geq 0 \\
    \mathcal{G}_{\text{vanilla VI}} &= \mathcal{G}_{\text{approx}}
    \end{aligned}
\end{equation}
The \textit{approximation gap} $\mathcal{G}_{\text{approx}}$ depends on the expressivity of the variational family $\mathcal{Q}$, specifically whether $\mathcal{Q}$ contains distributions arbitrarily close to the posterior --in the KL sense.
\citet{amortization_gap} demonstrate that, in the case of sample amortized inference, when the weights $\phi$ no longer are free but the output of an encoder $f \in \mathcal{F}$, inference cannot be better than in the non-sample-amortized case, and a positive \textit{amortization gap} is introduced:
\begin{equation}
\label{eq:sa}
    \begin{aligned}
    \mathcal{G}_{\text{sa}}(\mathcal{Q}, \mathcal{F}; \eta^*) &= \mathcal{G}_{\text{approx}}(\mathcal{Q}; f(\tX, \eta^*)) - \mathcal{G}_{\text{approx}}(\mathcal{Q}; \phi^*) \\
    &\geq 0 \\
    \mathcal{G}_{\text{sample amortized VI}} &= \mathcal{G}_{\text{approx}} + \mathcal{G}_{\text{sa}}
    \end{aligned}
\end{equation}
Where we denote as $\eta^*$ the optimal weights for the encoder $f$ inside the function family $\mathcal{F}$.
The gap terminology can be interpreted as follow: "theoretically, sample amortization cannot be beneficial in terms of KL divergence for the inference over a given sample $\tX$."

Using the same gap terminology, we can define gaps implied by our PAVI architecture.
Instead of picking the distribution $q$ inside the family $\mathcal{Q}$, consider picking $q$ from the \textit{plate-amortized} family $\mathcal{Q}_{\text{pa}}$ corresponding to $\mathcal{Q}$. Distributions in $\mathcal{Q}_{\text{pa}}$ are distributions from $\mathcal{Q}$ with the additional constraints that some weights have to be equal. Consequently, $\mathcal{Q}_{\text{pa}}$ is a subset of $\mathcal{Q}$:
\begin{equation}
    \mathcal{Q}_{\text{pa}} \subset \mathcal{Q}
\end{equation}
As such, looking for the optimal distribution inside $\mathcal{Q}_{\text{pa}}$ instead of inside $\mathcal{Q}$ cannot result in better performance, leading to a \textit{plate amortization gap}:
\begin{equation}
\label{eq:pa}
    \begin{aligned}
    \mathcal{G}_{\text{pa}}(\mathcal{Q}, \mathcal{Q}_{\text{pa}}; \psi^*, \phi^*) &=  \mathcal{G}_{\text{approx}}(\mathcal{Q}_{\text{pa}}; \psi^*) - \mathcal{G}_{\text{approx}}(\mathcal{Q}; \phi^*) \\
    &\geq 0 \\
    \mathcal{G}_{\text{PAVI-F}} &= \mathcal{G}_{\text{approx}} + \mathcal{G}_{\text{pa}}
    \end{aligned}
\end{equation}
Where we denote as $\psi^*$ the optimal weights for a variational distribution $q$ inside $\mathcal{Q}_{\text{pa}}$ --in the KL sense. The equation \ref{eq:pa} is valid for the PAVI-F scheme --see \cref{sec:encoding_schemes}. We can interpret it as follow: "theoretically, plate amortization cannot be beneficial in terms of KL divergence for the inference over a given sample $\tX$".

Now consider that encodings are no longer free parameters but the output of an encoder $f$. Similar to the case presented in \cref{eq:sa}, using an encoder cannot result in better performance, leading to an \textit{encoder gap}:
\begin{equation}
\label{eq:pa-e}
    \begin{aligned}
    \mathcal{G}_{\text{encoder}}(\mathcal{Q}_{\text{pa}}, \mathcal{F}; \psi^*, \eta^*) &=  \mathcal{G}_{\text{approx}}(\mathcal{Q}_{\text{pa}}; f(\tX, \eta^*)) - \mathcal{G}_{\text{approx}}(\mathcal{Q}_{\text{pa}}; \psi^*) \\
    &\geq 0 \\
    \mathcal{G}_{\text{PAVI-E}} &= \mathcal{G}_{\text{approx}} + \mathcal{G}_{\text{pa}} + \mathcal{G}_{\text{encoder}}
    \end{aligned}
\end{equation}
The equation $\cref{eq:pa-e}$ is valid for the PAVI-E scheme --see \cref{sec:encoding_schemes}.

The most complex case is the PAVI-E(sa) scheme, where we combine both plate and sample amortization.
Our argument cannot account for the resulting $\mathcal{G}_{\text{PAVI-E(sa)}}$ gap: both the PAVI-E and PAVI-E(sa) schemes rely upon the same encoder $f$.
In the PAVI-E scheme, $f$ is overfit over a dataset composed of the slices of a given data sample $\tX$.
In the PAVI-E(sa) scheme, the encoder is trained  over the whole distribution of the samples of the reduced model $\mathcal{M}^{\text{redu}}$.
Intuitively, it is likely that the performance of PAVI-E(sa) will always be dominated by the performance of PAVI-E, but --as far as we understand it-- the gap terminology cannot account for this discrepancy.

Comparing previous equations, we therefore have:
\begin{equation}
    \mathcal{G}_{\text{vanilla VI}} \leq \mathcal{G}_{\text{PAVI-F}} \leq \mathcal{G}_{\text{PAVI-E}}
\end{equation}
Note that those are \textit{theoretical} results, that do not necessarily pertain to optimization in practice.
In particular, in \cref{sec:exp_convergence_speed}\&\ref{sec:exp_scaling}, this theoretical performance loss is not observed empirically over the studied examples.
On the contrary, in practice our results can actually be better than non-amortized baselines, as is the case for the PAVI-F scheme in \cref{fig:scaling}.
We interpret this as a result of a simplified optimization problem due to plate amortization --with fewer parameters to optimize for, and mini-batching effects across different ground RVs.
A better framework to explain those discrepancies could be the one from \citet{bottou_tradeoffs_2007}: performance in practice is not only the reflection of an \textit{approximation error}, but also of an \textit{optimization error}.
A less expressive architecture --using plate amortization-- may in practice yield better performance.
Furthermore, for the experimenter, the theoretical gaps $\mathcal{G}_{\text{pa}}, \mathcal{G}_{\text{encoder}}$ are likely to be well "compensated for" by the lighter parameterization and faster convergence entitled by plate amortization.
\section{Supplemental results}
\subsection{GRE results sanity check}
\begin{figure}
    \centering
    \includegraphics[height=0.78\textheight]{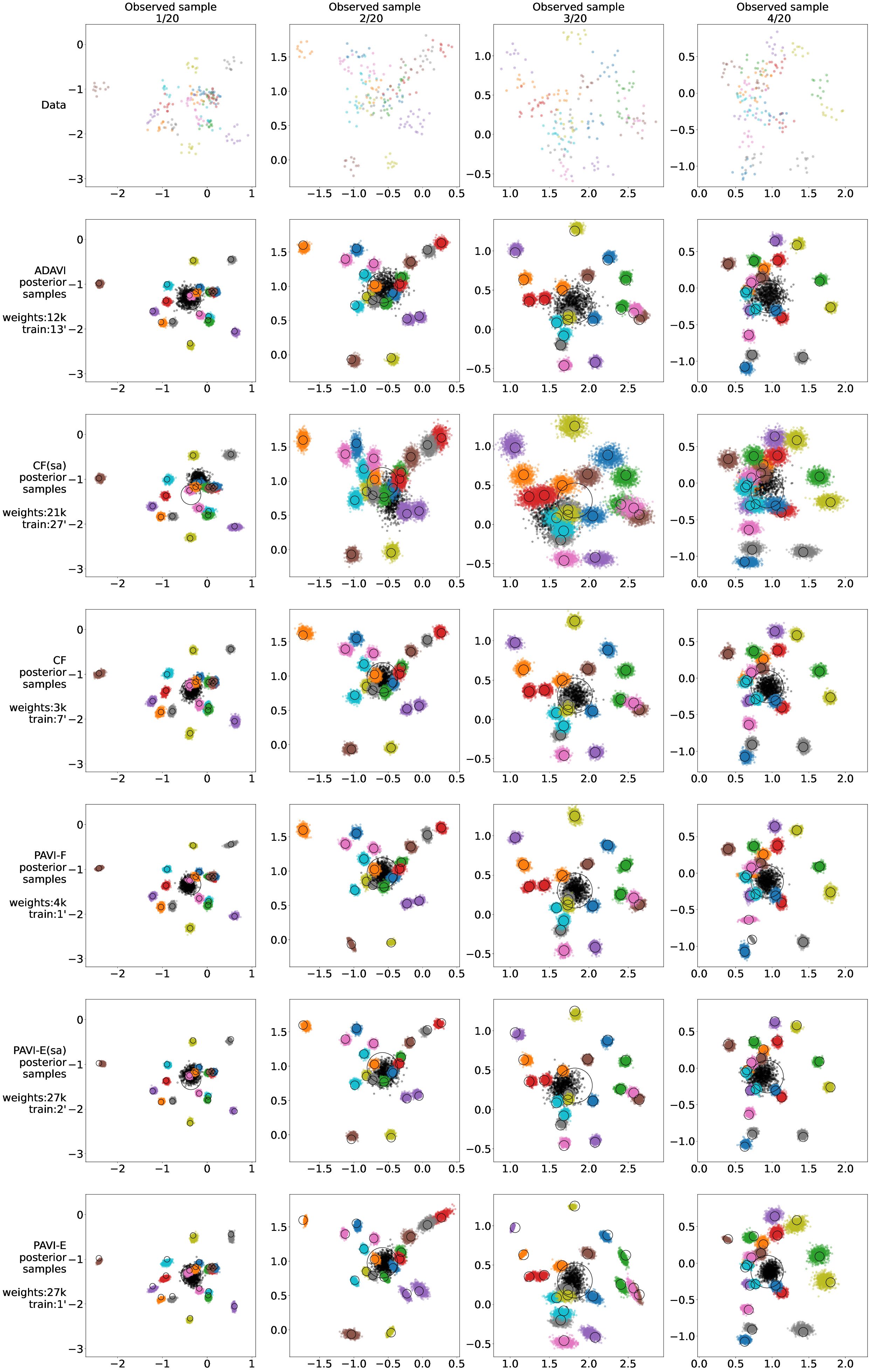}
    \caption{\textbf{GRE Sanity check} Inference methods present qualitatively correct results, making ELBO comparisons relevant in our experiments. \textit{On the topmost line}, we represent 4 different $\tX$ samples for the GRE model described in \cref{eq:GRE} with $\operatorname{Card}^\text{full}(\mathcal{P}_1) = 20$.
    Each set of colored points represent the $\tX_{n_1, \bullet}$ points belonging to one of the 20 groups.
    \textit{Bottom lines} represent the posterior samples for the methods used in \cref{sec:exp_scaling}.
    Colored points are sampled from the posterior of the groups means $\theta_1$, whereas black points are samples from the population mean $\theta_2$.
    We represent as black circles an analytical ground truth, centered on the correct posterior mean, and with a radius equal to 2 times the analytical posterior's standard deviation.
    \textbf{Correct posterior samples should be centered on the same point as the corresponding black circle, and 95\% of the points should fall within the black circle}.
    PAVI is represented on the 3 last lines, where we can observed a superior quality for the PAVI-F scheme, rivaling ADAVI and CF's performance with orders of magnitude less parameters and training time, as visible in \cref{fig:scaling}.}
    \label{fig:sanity}
\end{figure}

As exposed in the introduction of \cref{sec:exps}, in this work we focused on the usage of the ELBO as an inference performance metric \citep{blei_variational_2017}:
\begin{equation}
    \begin{aligned}
    \operatorname{ELBO}(q) &= \log p(X) - \operatorname{KL}(q(\Theta) || p(\Theta | X))
    \end{aligned}
\end{equation}
Given that the likelihood term $\log p(X)$ does not depend on the variational family $q$, differences in ELBOs directly transfer in differences in KL divergence, and provide with a straightforward metric to compare different variational posteriors.
Nonetheless, the ELBO doesn't provide with an absolute metric of quality.
As a sanity check, we want to assert the quality of the results presented in \cref{sec:exp_scaling} --that are transferable to \cref{sec:exp_convergence_speed}\&\ref{sec:exp_encoding_size}, based on the same model.
In \cref{fig:sanity} we plot the posterior samples of various methods against analytical ground truths, using the $\operatorname{Card}^\text{full}(\mathcal{P}_1) = 20$ case.
All the method's results are aligned with the analytical ground truth, with differences in ELBO translating meaningful qualitative differences in terms of inference quality.

\subsection{Experimental details - analytical examples}

All experiments were performed in Python, using the \textit{Tensorflow Probability} library \citep{dillon_tensorflow_2017}.
Throughout this section we refer to \textit{Masked Autoregressive Flows} \citep{papamakarios_masked_2018} as \textit{MAF}.
All experiments are performed using the Adam optimizer \citep{adam_kingma2014}.
At training, the ELBO was estimated using a Monte Carlo procedure with $8$ samples.
All architectures were evaluated over a fixed set of $20$ samples $\tX$, with 5 seeds per sample.
Non-sample-amortized architectures were trained and evaluated on each of those points.
Sample amortized architectures were trained over a dataset of $20,000$ samples separate from the 20 validation samples, then evaluated over the 20 validation samples.

\subsubsection{Plate amortization and convergence speed (\ref{sec:exp_convergence_speed})}

All 3 architectures (baseline, PAVI-F, PAVI-E) used:
\begin{itemize}
    \item for the flows $\mathcal{F}_i$, a MAF with $\left[ 32, 32 \right]$ hidden units;
    \item as encoding size, $128$
\end{itemize}
For the encoder $f$ in the PAVI-E scheme, we used a multi-head architecture with 4 heads of 32 units each, 2 ISAB blocks with 64 inducing points.

\subsubsection{Impact of encoding size (\ref{sec:exp_encoding_size})}

All architectures used:
\begin{itemize}
    \item for the flows $\mathcal{F}_i$, a MAF with $\left[ 32, 32 \right]$ hidden units, after an affine block with triangular scaling matrix.
    \item as encoding size, a value varying from $2$ to $16$
\end{itemize}

\subsubsection{Scaling with plate cardinalities (\ref{sec:exp_scaling})}

\textbf{ADAVI} \citep{ADAVI} we used:
\begin{itemize}
\item for the flows $\mathcal{F}_i$, a MAF with $\left[ 32, 32 \right]$ hidden units, after an affine block with triangular scaling matrix.
\item for the encoder, an encoding size of $8$ with a multi-head architecture with 2 heads of 4 units each, 2 ISAB blocks with 32 inducing points.
\end{itemize}

\textbf{Cascading Flows} \citep{CF} we used:
\begin{itemize}
    \item a mean-field distribution over the auxiliary variables $r$
    \item as auxiliary size, a fixed value of 8
    \item as flows, \textit{Highway Flows} as designed by the Cascading Flows authors
\end{itemize}

\textbf{PAVI-F} we used:
\begin{itemize}
    \item for the flows $\mathcal{F}_i$, a MAF with $\left[ 32, 32 \right]$ hidden units, after an affine block with triangular scaling matrix.
    \item an encoding size of 8
\end{itemize}

\textbf{PAVI-E} we used:
\begin{itemize}
    \item for the flows $\mathcal{F}_i$, a MAF with $\left[ 32, 32 \right]$ hidden units, after an affine block with triangular scaling matrix.
    \item for the encoder, an encoding size of $16$ with a multi-head architecture with 2 heads of 8 units each, 2 ISAB blocks with 64 inducing points.
\end{itemize}

\subsection{Details about our Neuroimaging experiment (\ref{sec:exp_fMRI})}

\subsubsection{Data description}
\label{sec:fMRI_data}
In this experiment we use data from the \textit{Human Connectome Project (HCP)} dataset \citep{HCP}.
We randomly select a cohort of $S=1,000$ subjects from this dataset, each subject being associated with $T=2$ resting state fMRI sessions \citep{rfMRI}.
We minimally pre-process the signal using the nilearn python library \citep{nilearn}:
\begin{enumerate}
    \item removing high variance confounds
    \item detrending the data
    \item band-filtering the data (0.01 to 0.1 Hz), with a repetition time of 0.74 seconds
    \item spatially smoothing the data with a 4mm Full-Width at Half Maximum
\end{enumerate}

For every subject, we extract the surface Blood Oxygenation Level Dependent (BOLD) signal of $N=314$ vertices corresponding to an average Broca's area \citep{heimEffectiveConnectivityLeft2009}.
We compare this signal with the extracted signal of $D=64$ DiFuMo components: a dictionary of brain spatial maps allowing for an effective fMRI dimensionality reduction \citep{difumo}.
Specifically, we compute the one-to-one Pearson's correlation coefficient of every vertex with every DiFuMo component.
The resulting connectome, with $S$ subjects, $T$ sessions, $N$ vertices and a connectivity signal with $D$ dimensions, is of shape $(S \times T \times N \times D)$.
We project this data --correlation coefficients lying in $\left] -1; 1 \right[$-- in an unbounded space using an inverse sigmoid function.

\subsubsection{Model description}
We use a model inspired from the work of \citet{kong_spatial_2018}.
We hypothesize that every vertex in Broca's area belongs to either one of $L=2$ functional networks.
This functional bi-partition would reflect the anatomical partition between \textit{pars opercularis} and \textit{pars triangularis} \citep{heimEffectiveConnectivityLeft2009, zhangConnectingConceptsBrain2020}.

Each network is a pattern of connectivity with the brain cortex, represented as a the correlation of the BOLD signal with the signal from the $D=64$ DiFuMo components.
We define $L=2$ such functional networks at the population level, that correspond to some "average" across the cohort of subjects.
Every subject has an individual connectivity, and therefore individual $L=2$ networks, that are considered as a Gaussian perturbation of the population networks, with variance $\epsilon$.
The connectivity of a given subject also evolves through time, giving rise to session-specific networks, that are a Gaussian perturbation of the subject networks with variance $\sigma$.
Finally, every vertex in Broca's area has its individual connectivity, and is a perturbation of one network's connectivity or the other's.
We model this last step as a Gaussian mixture distribution with variance $\kappa$.
We explicitly model the label $\operatorname{label}$ of a given vertex, and we consider this label constant across sessions.

The resulting model can be described as:
\begin{equation}
    \begin{aligned}
    S^\text{full}, T^\text{full}, N^\text{full}, D, L &= 1000, 2, 314, 64, 2 \\
    s^-, s^+ &= -6, 0 \\
    \substack{\forall l=1..L}: \quad \mu_{l} &\sim \operatorname{Uniform}(-4 \times \vec 1_D, 4 \times \vec 1_D) \\
    \substack{\forall l=1..L}: \quad \log \epsilon_{l} &\sim \operatorname{Uniform}(s^- \times \vec 1_D, s^+ \times \vec 1_D) \\
    \substack{\forall l=1..L\\ \forall s=1..S}: \quad \mu_{l, s} | \mu_l, \epsilon_l &\sim \mathcal{N}(\mu_l, \epsilon_l) \\
    \substack{\forall l=1..L}: \quad \log \sigma_{l} &\sim \operatorname{Uniform}(s^- \times \vec 1_D, s^+ \times \vec 1_D) \\
    \substack{\forall l=1..L\\ \forall s=1..S \\ \forall t=1..T}: \quad \mu_{l, s, t} | \mu_{l, s}, \sigma_l &\sim \mathcal{N}(\mu_{l, s}, \sigma_l) \\
    \substack{\forall l=1..L}: \quad \log \kappa_{l} &\sim \operatorname{Uniform}(s^- \times \vec 1_D, s^+ \times \vec 1_D) \\
    \substack{\forall s=1..S \\ \forall n=1..N}: \quad \operatorname{probs}_{s, n} &\sim \operatorname{Dirichlet}(1 \times \vec 1_L) \\
    \substack{\forall s=1..S \\ \forall n=1..N}: \quad \operatorname{label}_{s,n} | \operatorname{probs}_{s, n} &\sim \operatorname{Categorical}(\operatorname{probs}_{s,n}) \\
    \substack{\forall s=1..S \\ \forall t=1..T \\ \forall n=1..N}: \quad X_{s,t,n} | [\mu_{l,s,t}]_{l=1..L}, [\kappa_{l}]_{l=1..L}, \operatorname{label}_{s,n} &\sim \mathcal{N}(\mu_{\operatorname{label}_{s,n}, s, t}, \kappa_{\operatorname{label}_{s,n}})
    \end{aligned}
\end{equation}
The model contains 4 plates: the \textit{network} plate of full cardinality $L$ (that we did not exploit in our implementation), the \textit{subject} plate of full cardinality $S^\text{full}$, the \textit{session} plate of full cardinality $T^\text{full}$ and the \textit{vertex} plate of full cardinality $N^\text{full}$.

Our goal is to recover the posterior distribution of the networks $\mu$ --represented as networks in \cref{fig:broca}-- and the labels $\operatorname{label}$ --represented as the parcellation in \cref{fig:broca}-- given the observed connectome described in \cref{sec:fMRI_data}.

\subsubsection{PAVI implementation}

We used in this experiment the PAVI-F scheme, using:
\begin{itemize}
    \item for the RVs $\mu_l, \mu_{l,s}, \mu_{l,s,t}$:
    \begin{itemize}
        \item for the flows $\mathcal{F}_i$, a MAF with $\left[ 128, 128 \right]$ hidden units, following an affine block with diagonal scale
        \item for the encoding size: $128$
    \end{itemize}
    \item for the RVs $\epsilon_l, \sigma_l, \kappa_l, \operatorname{probs}_{s,n}, \operatorname{labels}_{s,n}$:
    \begin{itemize}
        \item for the flows $\mathcal{F}_i$, a MAF with $\left[ 8, 8 \right]$ hidden units, following an affine block with diagonal scale
        \item for the encoding size: $8$
    \end{itemize}
    \item for the reduced model, we used $S^\text{redu}=30$, $T^\text{redu}=1$ and $N^\text{redu}=32$.
\end{itemize}

To allow for the optimization over the discrete $\operatorname{label}_{s,n}$ RV, we used the Gumbell-Softmax trick, using a fixed temperature of $1.0$ \citep{gumbell_softmax, concrete}.

\section{Supplemental discussion}
\subsection{Plate amortization as a generalization of sample amortization}
In \cref{sec:plate_amortization} we introduced plate amortization as the application of the generic concept of amortization to the granularity of plates.
Taking a step back, there is actually an even stronger connection between sample amortization and plate amortization.
 
A HBM $p$ models the distribution of a given observed RV $X$ --jointly with the parameters $\Theta$.
Different samples $\tX_0, \tX_1, ...$ of the model $p$ are i.i.d. draws from the distribution $p(X)$.
$p$ can thus be considered as the model for "one sample".
Consider, instead of $p$, a "macro" model for the whole \textit{population} of samples one could draw from $p$.
The observed RV of that macro model would be the infinite collection of samples drawn from the same distribution $p(X)$.
In that light, the i.i.d. sampling of different $X$ values from $p$ could be interpreted as a plate of the macro model.
Thus, we could consider sample amortization as a instance of plate amortization for the "sample plate".
Or equivalently: plate amortization can be seen as the natural generalization of amortization beyond the particular case of sample amortization.

\subsection{Alternate formalism for SVI -- PAVI-E(sa) scheme}
In this work, we propose a different formalism for SVI, based around the concept of full HBM $\mathcal{M}^{\text{full}}$ versus reduced HBM $\mathcal{M}^{\text{redu}}$ sharing the same template $\mathcal{T}$.
This formalism is helpful to set up GPU-accelerated stochastic VI \citep{dillon_tensorflow_2017}, as it entitles a fixed computation graph -with the cardinality of the reduced model $\mathcal{M}^{\text{redu}}$- in which encodings are "plugged in" -either sliced from larger encoding arrays or as the output of an encoder applied to a data slice, see \cref{sec:encoding_schemes}\&\ref{sec:sharing_learning}.
Particularly, our formalism doesn't entitle a control flow over models and distributions, which can be hurtful in the context of \textit{compiled} computation graphs such as in \textit{Tensorflow} \citep{tensorflow2015-whitepaper}.

The reduced model formalism is also meaningful in the PAVI-E(sa), where we train and amortized variational posterior over $\mathcal{M}^{\text{redu}}$ and obtain "for free" a variational posterior for the full model $\mathcal{M}^{\text{full}}$ --see \cref{sec:sharing_learning}.
In this context, our scheme is no longer a different take on hierarchical, batched SVI: the cardinality of the full model is truly independent from the cardinality of the training, and is only simulated as a scaling factor in the stochastic training --see \cref{sec:stochastic_training}.
We have the intuition that fruitful research directions could stem from this concept.

\subsection{Benefiting from structure in inference}
Conceptually, all our contributions can be abstracted through the notion of plate amortization -see \cref{sec:plate_amortization}.
Plate amortization is particularly useful in the context of heavily parameterized density approximators such as normalizing flows, but is not tied to it: plate-amortized Mean Field \citep{blei_variational_2017} or ASVI \citep{ASVI} schemes are also possible to use.
Plate amortization can be viewed as the amortization of common density approximators across different sub-structures of a problem.
This general concept could have applications in other highly-structured problem classes such as graphs or sequences \citep{wu_comprehensive_2020, rnn_review}.

\subsection{Towards user-friendly Variational Inference}
By re-purposing the concept of amortization at the plate level, our goal is to propose clear computation versus precision trade-offs in VI.
Hyper-parameters such as the encoding size --as illustrated in \cref{fig:encoding_size} (right)-- allow to clearly trade inference quality in exchanged for a reduced memory footprint.
On the contrary, in classical VI, changing $\mathcal{Q}$'s parametric form --for instance switching from Gaussian to Student distributions-- can have a strong and complex impact both on number of weights and inference quality \citep{blei_variational_2017}.
By allowing the usage of normalizing flows in very large cardinality regimes, our contribution aims at de-correlating approximation power and computational feasibility.
In particular, having access to expressive density approximators for the posterior can help experimenters diversify the proposed HBMs, removing the need of properties such as conjugacy to obtain meaningful inference \citep{Gelman_book}.
Combining clear hyper-parameters and scalable yet universal density approximators, we tend towards a user-friendly methodology in the context of large population studies VI.

\end{document}